\def\year{2018}\relax
\documentclass[letterpaper]{article} 
\usepackage{aaai18}  
\usepackage{times}  
\usepackage{helvet}  
\usepackage{courier}  
\usepackage{url}  
\usepackage{graphicx}  
\usepackage{amsfonts}
\usepackage{amstext}
\usepackage{amsmath}
\usepackage{algorithm}
\usepackage{algorithmic}
\usepackage{amssymb}
\usepackage{bm}
\usepackage{amsmath}
\usepackage{subfigure}

\begin{document}

%
\title{Generative Adversarial Mapping Networks}
%
\author{Jianbo Guo\textsuperscript{*}, Guangxiang Zhu\textsuperscript{*}, Jian Li\\
	Institute for Interdisciplinary
	Information Sciences, Tsinghua University, Beijing,
	China\\
	jianboguo@outlook.com, insmileworld@gmail.com, lijian83@mail.tsinghua.edu.cn\\
}
\maketitle
\begin{abstract}
Generative Adversarial Networks (GANs) have shown impressive performance in generating
photo-realistic images. They fit generative models by minimizing certain distance measure
between the real image distribution and the generated data distribution. Several distance measures have been used, such as Jensen-Shannon divergence, $f$-divergence, and Wasserstein distance, and choosing an appropriate distance measure is very important for training the generative network. In this paper, we choose to use
the maximum mean discrepancy (MMD) as the distance metric, which has several
nice theoretical guarantees.
In fact, generative moment matching network (GMMN) \cite{li2015generative}
is such a generative model
which contains only one generator network $G$ trained by directly minimizing MMD between the real and generated distributions.
However, it fails to generate meaningful samples on challenging benchmark datasets, such as CIFAR-10 and LSUN.
To improve on GMMN, we propose to add an extra network $F$, called mapper.
$F$ maps both real data distribution and generated data distribution from the original data space to a feature representation space $\mathcal{R}$, and
it is trained to maximize MMD between the two mapped distributions in $\mathcal{R}$,
while the generator $G$ tries to minimize the MMD.
We call the new model \textit{generative adversarial mapping networks} (GAMNs).
We demonstrate that the adversarial mapper $F$ can help $G$ to better capture the underlying data distribution. We also show that GAMN significantly outperforms GMMN, and is also superior to or comparable with other state-of-the-art GAN based methods on MNIST, CIFAR-10 and LSUN-Bedrooms datasets.
\end{abstract}

\section{Introduction}
Generative adversarial networks (GANs) \cite{goodfellow2014generative}
have attracted much attentions recently due to their capability of capturing the underlying real data distribution $\mathbb{P}_r$ and synthesizing new samples.
GANs typically consist of two networks, a generator $G$ and a discriminator $D$.
The generator $G$ takes a random variable $z$ sampled from a prior distribution $p(z)$
(e.g., uniform or normal) and outputs a generated sample (e.g., an image) of interest $y=G(z)$ via a feedforward neural network.
Hence, the generator defines the generated data distribution $\mathbb{P}_g$ implicitly: $y=G(z), z\sim p(z)$.
The basic idea behind GANs is to train $G$ and $D$ simultaneously: $D$ is trained to distinguish real data samples from fake samples generated by $G$ while $G$ is trained to fool $D$. GANs' training process is like a two-player game, and the global equilibrium achieves if and only if $\mathbb{P}_g=\mathbb{P}_r$ \cite{goodfellow2014generative}.
Following this, several variants of GAN have been proposed to
minimize different probability distances/divergences
between the real data distribution and the generated distribution,
such as $f$-divergences \cite{nowozin2016f} and Wasserstein distance \cite{arjovsky2017wasserstein,gulrajani2017improved} between $\mathbb{P}_r$ and $\mathbb{P}_g$. Some of those variants utilize $D$ to
estimate the distance between $\mathbb{P}_r$ and $\mathbb{P}_g$
(even though $D$ is not explicitly trained for classification,
it can still be thought as an implicit discriminator).

\citeauthor{li2015generative} propose a generative model, called
generative moment matching network (GMMN).
It includes only one network, i.e. a generator $G$, and uses
the maximum mean discrepancy (MMD) \cite{gretton2012kernel} to determine the distance between $\mathbb{P}_r$ and $\mathbb{P}_g$.
MMD has several desirable theoretical guarantees over other distance measures.
For example, it admits an efficient unbiased estimator (see e.g., \cite{gretton2012kernel}).
\footnote{
	On the other hand, we take Wasserstein distance for example.
	It can be written as
	$W(P,Q)=\sup_{f\text{ is 1-Lipschitz}} \mathbb{E}[\int f d(P-Q)]$. However,
	the empirical estimator according to the formula
	can be a very biased estimation for $W(P,Q)$ in high dimensions, unless the number of samples is exponential (see e.g.,\cite{arora2017generalization}).	
	}
GMMN can be trained by MMD distance minimization to learn the underlying data distribution $\mathbb{P}_r$. To further boost the performance of GMMN, \citeauthor{li2015generative} introduce an auto-encoder to GMMN, which is referred to as GMMN+AE. They first train an auto-encoder network and produce the code representation of real data, then fix the auto-encoder and apply a GMMN to learn the code distribution by minimizing MMD between data code and generated code. For real data generation, they first use the learned GMMN to yield a code sample and then pass it to the decoder of the previously fixed auto-encoder to generate sample in the real data space.
Both GMMN and GMMN+AE work well on MNIST \cite{lecun1998gradient} and the Toronto Face Database\cite{susskind2010toronto}. However, they both fail to generate meaningful samples on more challenging datasets like CIFAR-10 \cite{krizhevsky2009learning} and LSUN-Bedrooms dataset \cite{yu2015lsun} which contain images with much more complex structures and contents. Besides, the batch size required to train the GMMN and GMMN+AE is too large ($1000$ in the original paper).

In this paper, we integrate the adversarial training framework of GAN with the advantage of
MMD distance, to further improve the generative model.
More concretely, we add an extra network $F$, called mapper,
to replace the auto-encoder in GMMN+AE.
The mapper $F$ maps both $\mathbb{P}_r$ and $\mathbb{P}_g$ from the real data space to a feature representation space $\mathcal{R}$. Its functionality is somewhat
similar to that of the auto-encoder in GMMN+AE:
we would like to work in the representation space, which is relatively low-dimensional and
thus easier for estimating the distribution distance.
However, our training processes are entirely different: The auto-encoder is trained separately from the generator $G$ (the only one network in GMMN) and will be fixed afterwards. In comparison, the mapper $F$ in our model
is trained with $G$ simultaneously like in GANs:
$F$ aims to maximize MMD between the two mapped distributions in $\mathcal{R}$ and $G$ tries to minimize it.
As we will demonstrate in our experiments,
this simple change can yield significant improvement over GMMN and GMMN+AE.
We name our new model Generative Adversarial Mapping Networks (GAMNs).

We summarize our main contributions as follows:
\begin{itemize}
	\item We propose a new generative model, called GAMN, based on MMD distance and
	the adversarial training idea in GAN.
	Our model is quite simple, yet effective in generating realistic
	images and fairly stable in training.
	
	\item On toy datasets, e.g., a mixture of $8$ Gaussians, a mixture of $25$ Gaussians, and Swiss Roll, we show that GAMN can learn the underlying distribution better than the
	state-of-the-art GAN based methods such as WGAN \cite{arjovsky2017wasserstein} and improved WGAN \cite{gulrajani2017improved}.
	
	\item On MINIST, CIFRA-10 and LSUN-Bedrooms dataset, GAMN can produce images of high quality which are significantly better than the generated samples by GMMN and GMMN+AE and better than or comparable with those by the state-of-the-art GAN based methods like WGAN and improved WGAN. In addition, the batch size required for training is also much smaller than that of GMMN and GMMN+AE.
\end{itemize}

\section{Preliminary}

\subsection{GAN Framwork}
GANs \cite{goodfellow2014generative} define two networks,
the generator $G$ and the discriminator $D$. $G$ is a neural network which takes a random input vector $z$ sampled from a fixed prior distribution $p(z)$ (e.g., uniform or normal) and maps it to a sample of interest $G(z)$ in the real data space. $D$ is another neural network that takes a real sample or a generated fake sample by $G$ as input, and attempts to distinguish between them. Both networks are trained to outwit each other, that is, $G$ is trained to generate "real" enough samples to confuse $F$, and $F$ is required to tell the real samples from the generated samples. Mathematically, training GANs is to do the following minmax optimization:
\begin{equation}
\min_G \max_D \mathbb{E}_{ x\sim \mathbb{P}_r}\left[\log\left(D(x)\right)\right] + \mathbb{E}_{ y\sim \mathbb{P}_g}\left[\log\left(1-D(y)\right)\right]
\end{equation}
where $\mathbb{P}_r$ is the real data distribution and $\mathbb{P}_g$ is the generated data distribution. $\mathbb{P}_g$ is defined by $G$ and $p(z)$ implicitly: $y=G(z)$, $z\sim p(z)$.

There are some variants of GANs where the minmax optimization is to minimize some other distance between $\mathbb{P}_r$ and $\mathbb{P}_g$. Take WGAN \cite{arjovsky2017wasserstein} for example. WGAN tries to do the following Wasserstein distance minimization:
\begin{equation}
\min_G \max_D \mathbb{E}_{x\sim \mathbb{P}_r}\left[D(x)\right] - \mathbb{E}_{y\sim \mathbb{P}_g}\left[D(y)\right]
\end{equation}
where the mapping induced by $D$ is required to be 1-Lipschitz.
Here, training the discriminator here is to obtain estimate of Wasserstein distance between the two distributions and training the generator is to minimize it.

\subsection{Maximum Mean Discrepancy}
Maximum Mean Discrepancy (MMD) \cite{gretton2012kernel} is a test statistic to determine if two samples are drawn from different distributions, defined by the largest difference in expectations over functions in the unit ball of a reproducing kernel Hilbert space $\mathcal{H}$ associated with a kernel $k(\cdot, \cdot)$. Formally, MMD between the real data distribution $\mathbb{P}_r$ and the generated data distribution $\mathbb{P}_g$ is defined by
$$
L_{\text{MMD}}(\mathbb{P}_r, \mathbb{P}_g) = \sup_{\lVert f \lVert_\mathcal{H}\le 1} \mathbb{E}_{x\sim \mathbb{P}_r} \left[f(x)\right] - \mathbb{E}_{y\sim \mathbb{P}_g} \left[f(y)\right]
$$
By kernel tricks, one can obtain
\begin{equation}
\begin{aligned}
L_{\text{MMD}}(\mathbb{P}_r, \mathbb{P}_g) &= \Big(\mathop{\mathbb{E}}\limits_{x, x'\sim p} \left[k(x, x')\right] - 2\mathop{\mathbb{E}}\limits_{x\sim p, y\sim q}\left[k(x, y)\right] \\
&+ \mathop{\mathbb{E}}\limits_{y, y'\sim p}\left[k(y, y')\right]\Big)^{\frac{1}{2}}
\end{aligned}
\end{equation}
which can be estimated by
\footnote{It is very close to the unbiased empirical estimator of MMD mentioned in \cite{gretton2012kernel}.} \cite{li2015generative}
\begin{equation}\label{estimator}
\begin{aligned}
\hat{L}_{\text{MMD}}(X, Y) &= \frac{1}{m^2}\Big(\sum_{i, i'} k(x_i, x_{i'}) - 2\sum_{i,j}k(x_i, y_j) \\
&+ \sum_{j, j'} k(y_j, y_{j'})\Big)^{\frac{1}{2}}
\end{aligned}
\end{equation}
where $X:=\{x_i\}_{i=1}^m$ and $Y:=\{y_j\}_{j=1}^m$ are independently and identically sampled from $\mathbb{P}_r$ and $\mathbb{P}_g$ respectively.

It is known that with a Gaussian kernel $k$, $\mathbb{P}_r=\mathbb{P}_g$ if and only if $L_{\text{MMD}}(\mathbb{P}_r, \mathbb{P}_g)=0$.
To train a generator to make $\mathbb{P}_g$ approximate $\mathbb{P}_r$, GMMN minimizes MMD between real distribution and generated distribution directly
\begin{equation}
\min_G L_{\text{MMD}}(\mathbb{P}_r, \mathbb{P}_g)
\end{equation}
where a mixture of $K$ Gaussian kernels is used: $k(x, x') = \sum_{q=1}^{K}k_{\sigma_q}(x, x')$ and $k_{\sigma_q}(x, x')$ is a Gaussian kernel with bandwidth $\sigma_q$. We denote $k_{\bm{\sigma}}$ as the mixture of Gaussian kernels for later use, where $\bm{\sigma}$ is the bandwidth hyperparameters. This method can generate good samples on MNIST and the Toronto Face Dataset.

\section{Generative Adversarial Mapping Networks}
Directly minimizing MMD between  $\mathbb{P}_r$ and $\mathbb{P}_g$ with Gaussian kernels, GMMN fails to generate meaningful samples on more challenging datasets such as CIFAR-10 and LSUN-Bedrooms dataset. By Taylor expansion, minimizing MMD is equivalent to minimizing a distance between all moments of the two distributions \cite{li2015generative}. We think this simple moment matching cannot capture the underlying data distribution of natural images with complex inner structures and spatial relation. Inspired by GANs, we introduce an extra network $F$, called mapper, to GMMN to help the generator $G$ model the underly distribution better. $F$ maps both real data distribution $\mathbb{P}_r$ and generate data distribution $\mathbb{P}_g$ from the original data space to a feature representation space $\mathcal{R}$. We denote $F(\mathbb{P}_r)$ and $F(\mathbb{P}_g)$ as the mapped real distribution and mapped generated distribution respectively. For natural image generation, we set $F$ to be
a convolutional neural network (CNN) \cite{lecun1995convolutional,krizhevsky2012imagenet} to learn hierarchical feature representations for images \cite{kavukcuoglu2010learning}.
Then we perform the MMD minimization between the two mapped distributions $F(\mathbb{P}_r)$
and $F(\mathbb{P}_g)$. Mathematically,
\begin{equation*}
\min_G L_{\text{MMD}}\left(F(\mathbb{P}_r), F(\mathbb{P}_g)\right)
\end{equation*}
where $F(\mathbb{P}_*)$ is implicitly defined by $\tilde{x}\sim F(\mathbb{P}_*) \Leftrightarrow \tilde{x}=F(x), x\sim \mathbb{P}_*$. Since we have learned the image feature representations by the mapper $F$, MMD with Gaussian kernels is enough now. In the following, we refer to $L_{\text{MMD}}(\cdot, \cdot)$ as the
MMD distance between two distributions with a mixture of Gaussian kernels $k_{\bm{\sigma}}$ (this distance is also used in GMMNs \cite{li2017mmd}).

What kind of mapper $F$ do we need? Intuitively, $L_{\text{MMD}}(\mathbb{P}_r, \mathbb{P}_g)=0$ implies $L_{\text{MMD}}\left(F(\mathbb{P}_r), F(\mathbb{P}_g)\right)=0, \forall F$.  Thus we need the largest MMD between the two mapped distributions to be small enough if we want $\mathbb{P}_g$ to approximate $\mathbb{P}_r$. Following from this reasoning, we obtain the following minmax optimization:
\begin{equation}\label{no-reg}
\min_G \max_{F\in\mathcal{F}}  L_{\text{MMD}}\left(F(\mathbb{P}_r), F(\mathbb{P}_g)\right)
\end{equation}
to make $\mathbb{P}_g$ to approximate $\mathbb{P}_r$. $\mathcal{F}$ is the set of candidate functions. From Equation (\ref{no-reg}), $F$ is trained to maximize MMD between the mapped distributions while $G$ is trained to minimize it. We call our model generative adversarial mapping network (GAMN) due to the adversarial mapping in the above optimization.

\begin{algorithm}[t]
	\caption{Learning algorithm for GAMN. }
	\label{alg}
	\hspace*{0.02in} {\bf Require:}
	Regularization term Reg, regularization strength $\lambda$, the number of iterations to train the mapper per round $n_{mapper}$, the number of iterations to train the generator per round $n_{generator}$, the batch size $m$, Adam hyperparameters $\beta_1, \beta_2$, the learning rate $\alpha$, bandwidth hyperparameters $\bm{\sigma}$.\\ %
	\hspace*{0.02in} {\bf Require:}
	Initial mapper parameters $w_0$, initial generator parameters $\theta_0$.
	\begin{algorithmic}[1]
		\WHILE{$\theta$ has not converged}
		\FOR{ $t = 1, \dots, n_{mapper} $}
		\STATE Sample $X:=\{x_i\}_{i=1}^m \sim \mathbb{P}_r$ a batch from the real data.
		\STATE Sample random vectors $\{z_i\}_{i=1}^m \sim p(z)$.
		\STATE $y_i = G_{\theta}(z_i), \forall i\in[m] $; $Y:=\{y_i\}_{i=1}^m$.
		\STATE $\begin{aligned}
		g_w \leftarrow &\nabla_w \hat{L}_{\text{MMD}}(F_w(X), F_w(Y)) +\lambda \nabla_w\text{Reg} \\
		\end{aligned}$
		\STATE $w \leftarrow w + \text{Adam}(g_w, \alpha, \beta_1, \beta_2)$
		\ENDFOR
		\FOR{ $t = 1, \dots, n_{generator} $}
		\STATE Sample $X:=\{x_i\}_{i=1}^m \sim \mathbb{P}_r$ a batch from the real data.
		\STATE Sample random vectors $\{z_i\}_{i=1}^m \sim p(z)$.
		\STATE $y_i = G_{\theta}(z_i), \forall i\in[m] $; $Y:=\{y_i\}_{i=1}^m$.
		\STATE $g_\theta \leftarrow \nabla_\theta \hat{L}_{\text{MMD}}(F_w(X), F_w(Y))$ 
		\STATE $\theta \leftarrow \theta - \text{Adam}(g_\theta, \alpha, \beta_1, \beta_2)$
		\ENDFOR
		\ENDWHILE
	\end{algorithmic}
\end{algorithm}
However, we notice that $\max_{F\in\mathcal{F}}  L_{\text{MMD}}\left(F(\mathbb{P}_r), F(\mathbb{P}_g)\right)$ may not be bounded since we can make $L_{\text{MMD}}\left(F(\mathbb{P}_r), F(\mathbb{P}_g)\right)$ infinity by multiplying $F$ by any large enough factor. To resolve this, some regularization is needed. In this paper, we restrict $\sum_{i=1}^d F(\cdot)[i]$ to be $1-$Lipschitz, where $F(\cdot)[i]$ is the $i^{th}$ component of function $F(\cdot)$ and $d$ is the number of dimensions of the feature representation space $\mathcal{R}$. By using the gradient penalty regularization proposed by \citeauthor{gulrajani2017improved}, training GAMN becomes
\begin{equation}\label{gp-reg}
\begin{aligned}
\min_G \max_{F}  L_{\text{MMD}}\left(F(\mathbb{P}_r), F(\mathbb{P}_g)\right) + \lambda\text{GP}
\end{aligned}
\end{equation}
where $\lambda$ is the regularization strength and $ \text{GP}=\mathbb{E}_{\hat{x}\sim \mathbb{P}_{\hat{x}}}\left[\left( \left\|\nabla_{\hat{x}} \left(\sum_{i=1}^d F(\hat{x})[i]\right)\right\|_2 -1 \right)^2\right]$. $\mathbb{P}_{\hat{x}}$ is defined implicitly by sampling uniformly along straight lines between pairs of points sampled from the data distribution $\mathbb{P}_r$ and the generator distribution $\mathbb{P}_g$. Note that the regularization term is only used when training the mapper $F$. We also find both $L1$  and $L2$ regularizations work well, especially on the parameters in normalization layers of $F$ only, because the parameters in normalization layers determine the scaling of the normalized output before the non-linear activation function. Formally, denote $w_{n}^F$ as the parameters in normalization layers of $F$, and then Equation (\ref{no-reg}) with $L1$ regularization and $L2$ regularization are
\begin{equation}\label{l2-reg}
\min_G \max_{F}  L_{\text{MMD}}\left(F(\mathbb{P}_r), F(\mathbb{P}_g)\right) + \lambda \lVert w_{n}^F \lVert_2^2
\end{equation}
and
\begin{equation}\label{l1-reg}
\min_G \max_{F}  L_{\text{MMD}}\left(F(\mathbb{P}_r), F(\mathbb{P}_g)\right) + \lambda \lVert w_{n}^F \lVert_1
\end{equation}
respectively.

We use a neural network parameterized with weight $\theta$ as the generator $G_{\theta}$ to produce samples from random vectors and another neural network parameterized with weight $w$ as the mapper $F_w$ to map both $\mathbb{P}_r$ and $\mathbb{P}_g$ to the space $\mathcal{R}$. Now for simplicity we rewrite Equation (\ref{gp-reg}-\ref{l1-reg}) as
\begin{figure*}[!h]
	\centering
	\begin{tabular}{llll}
		\centering
		
		\hspace{30pt}
		\textbf{WGAN} &\hspace{15pt}
		\textbf{improved WGAN} & \hspace{35pt}
		\textbf{GMMN} & \hspace{35pt}
		\textbf{GAMN} \\
		\hline
		\vspace{-7pt}
		\\
		\textbf{8 Gaussians} &&&
		\vspace{1pt}
		\\
		\hspace{-8pt}
		\begin{minipage}{0.2\linewidth}
			\includegraphics[width=\textwidth]{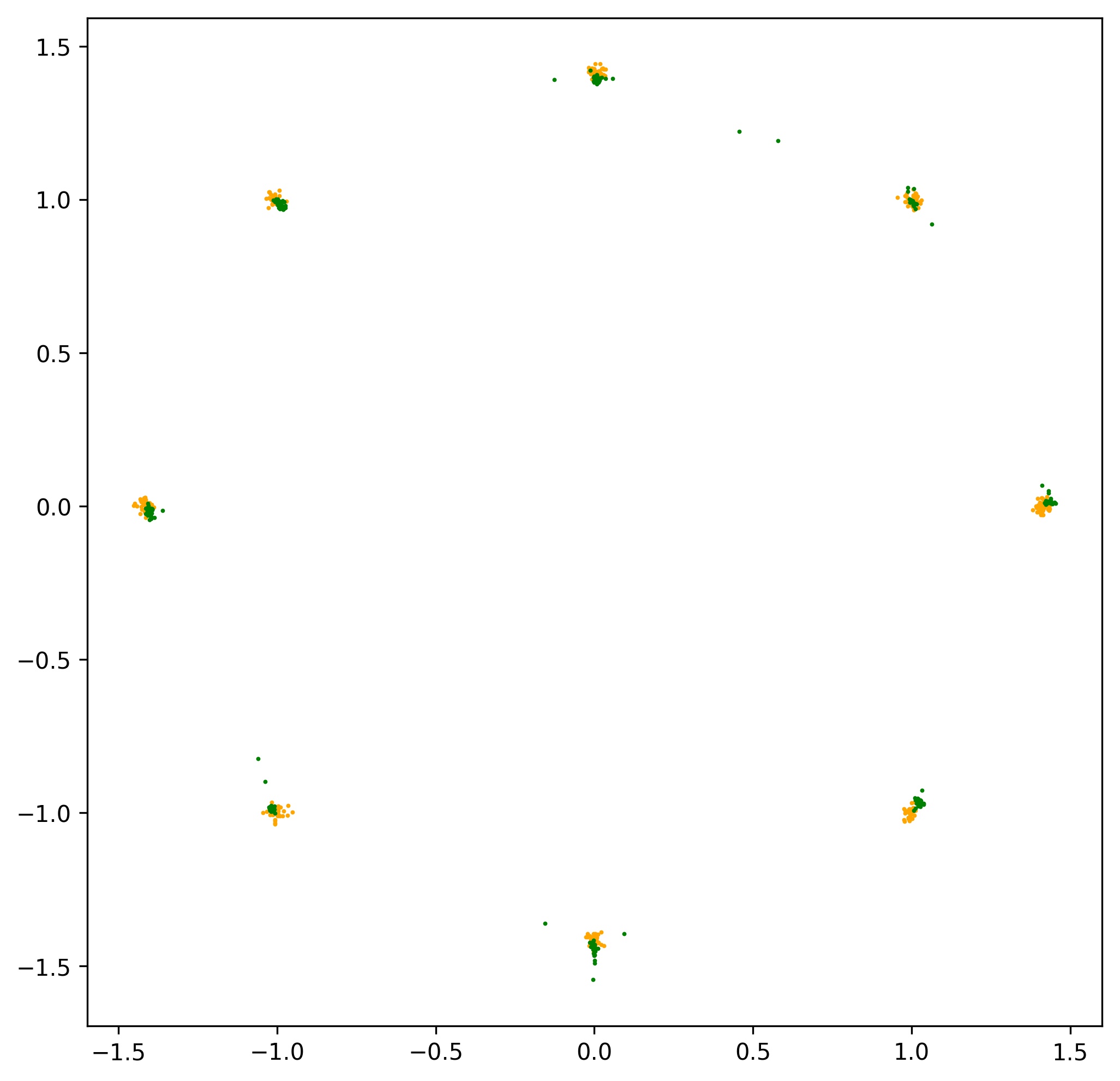}
		\end{minipage}
		&
		\hspace{-14pt}
		\begin{minipage}{0.2\linewidth}
			\includegraphics[width=\textwidth]{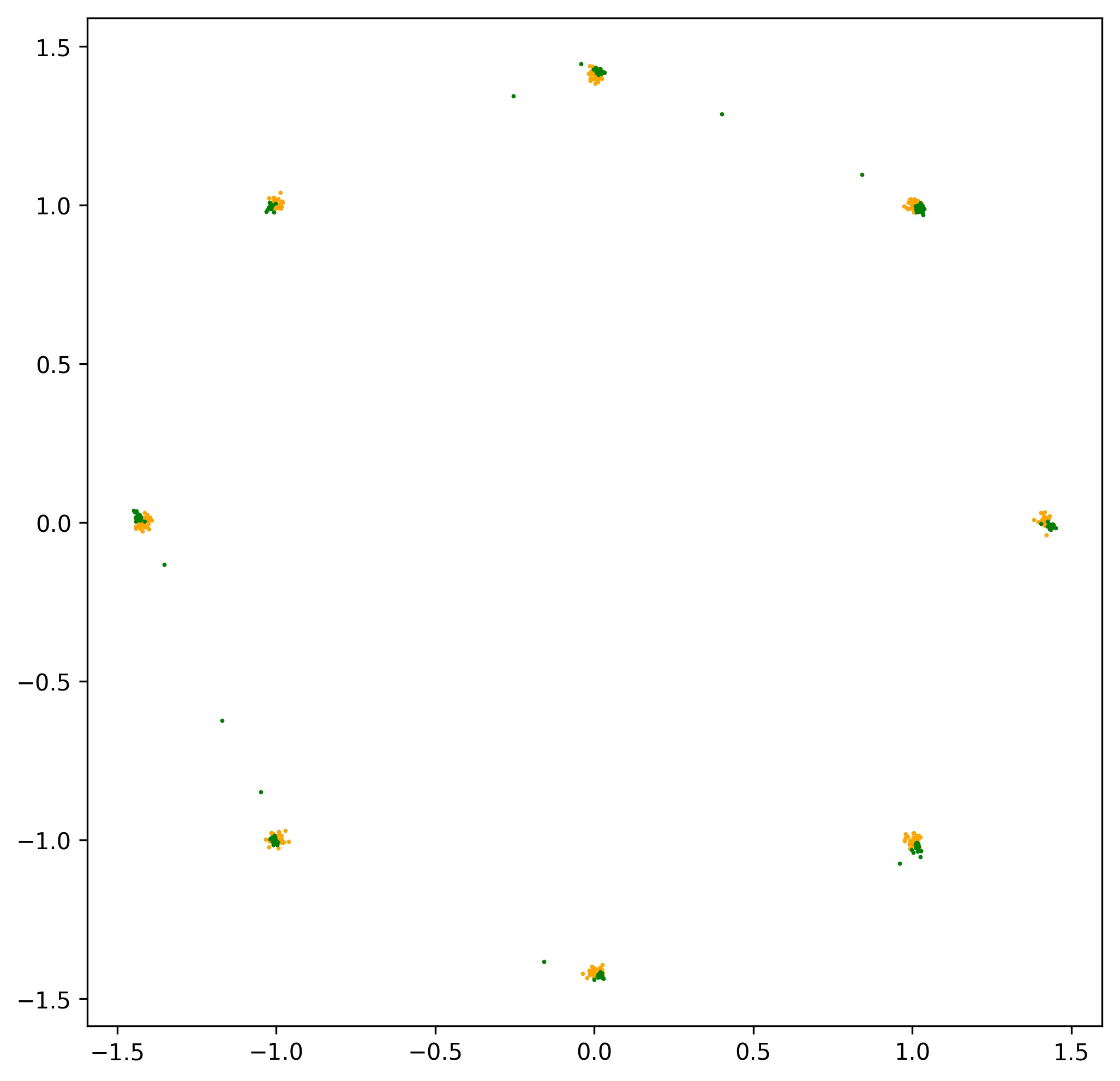}
		\end{minipage}
		&
		\hspace{-14pt}
		\begin{minipage}{0.2\linewidth}
			\includegraphics[width=\textwidth]{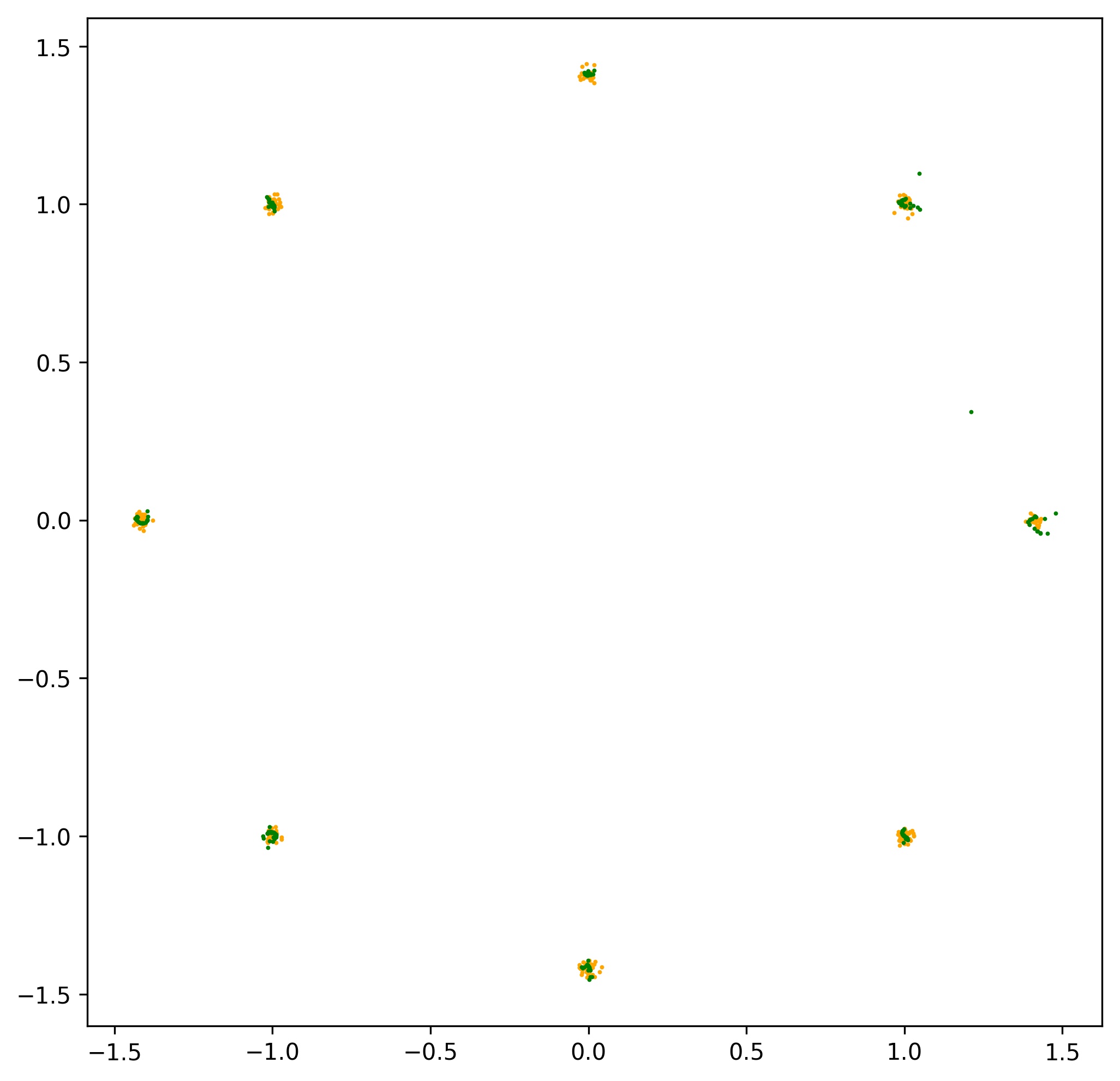}
		\end{minipage}
		&
		\hspace{-14pt}
		\begin{minipage}{0.2\linewidth}
			\includegraphics[width=\textwidth]{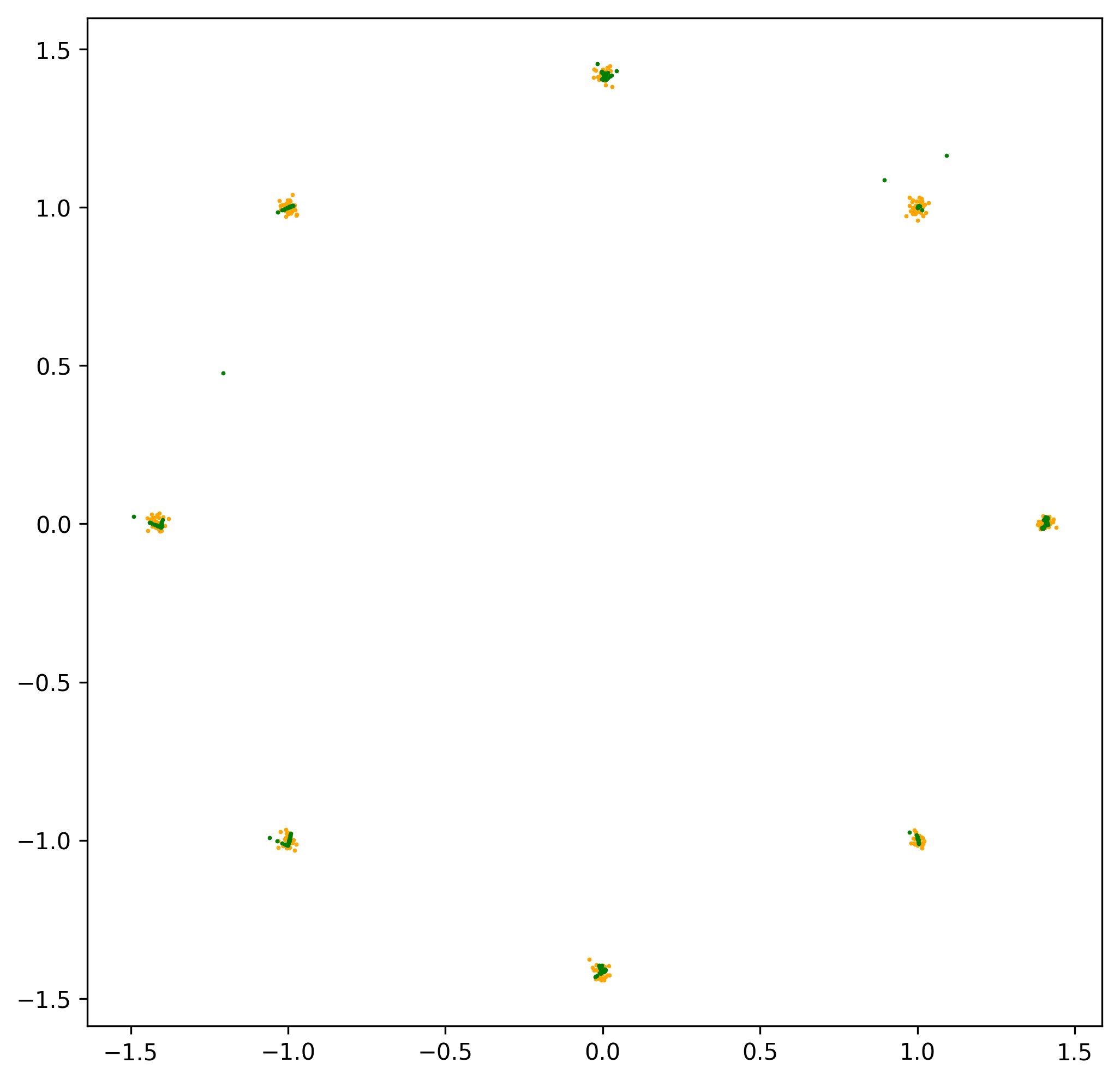}
		\end{minipage}
		\vspace{2pt}
		\\
		\textbf{25 Gaussians} &&&
		\vspace{1pt}
		\\
		\hspace{-8pt}
		\begin{minipage}{0.2\linewidth}
			\includegraphics[width=\textwidth]{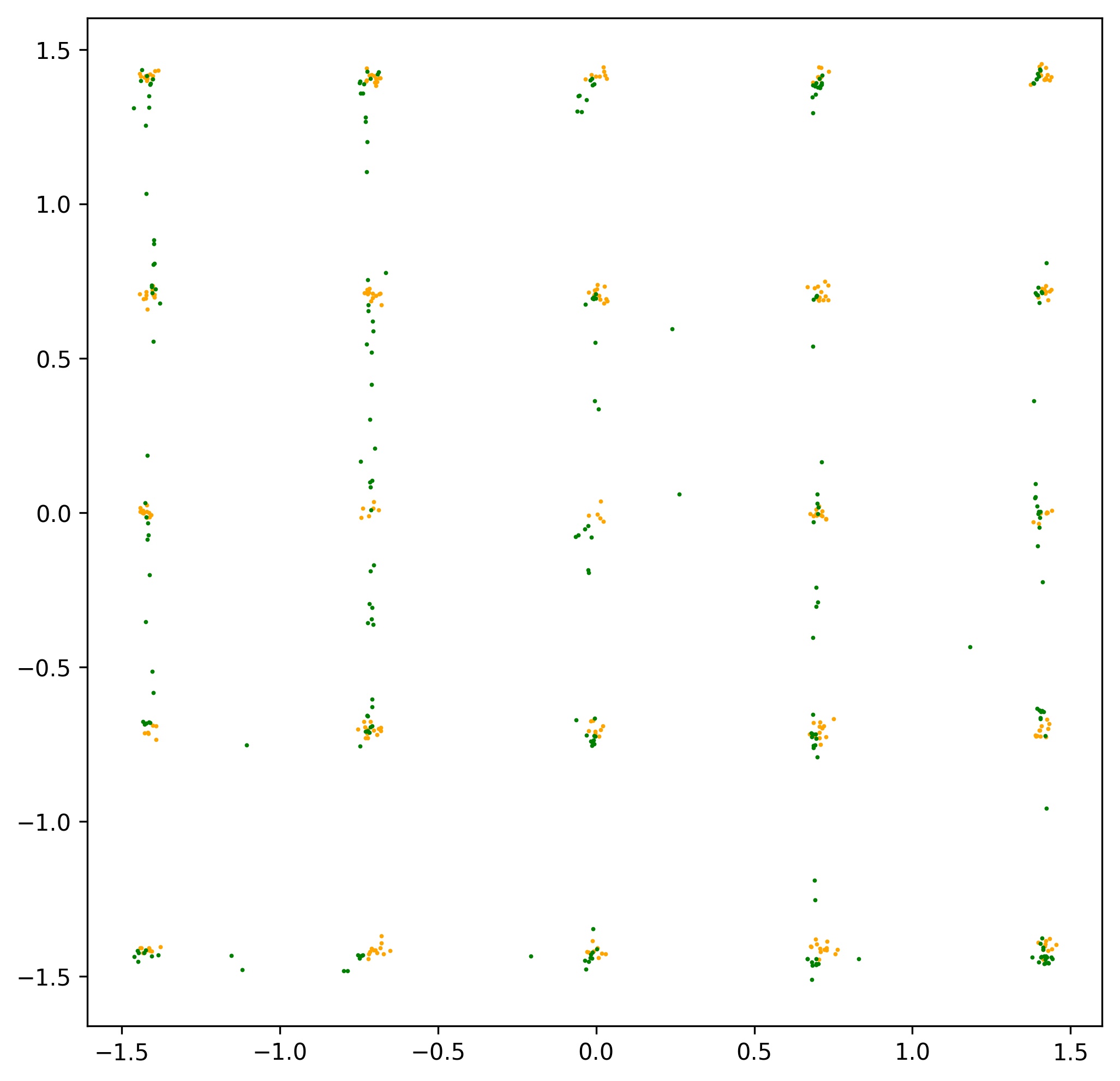}
		\end{minipage}
		&
		\hspace{-14pt}
		
		\begin{minipage}{0.2\linewidth}
			\includegraphics[width=\textwidth]{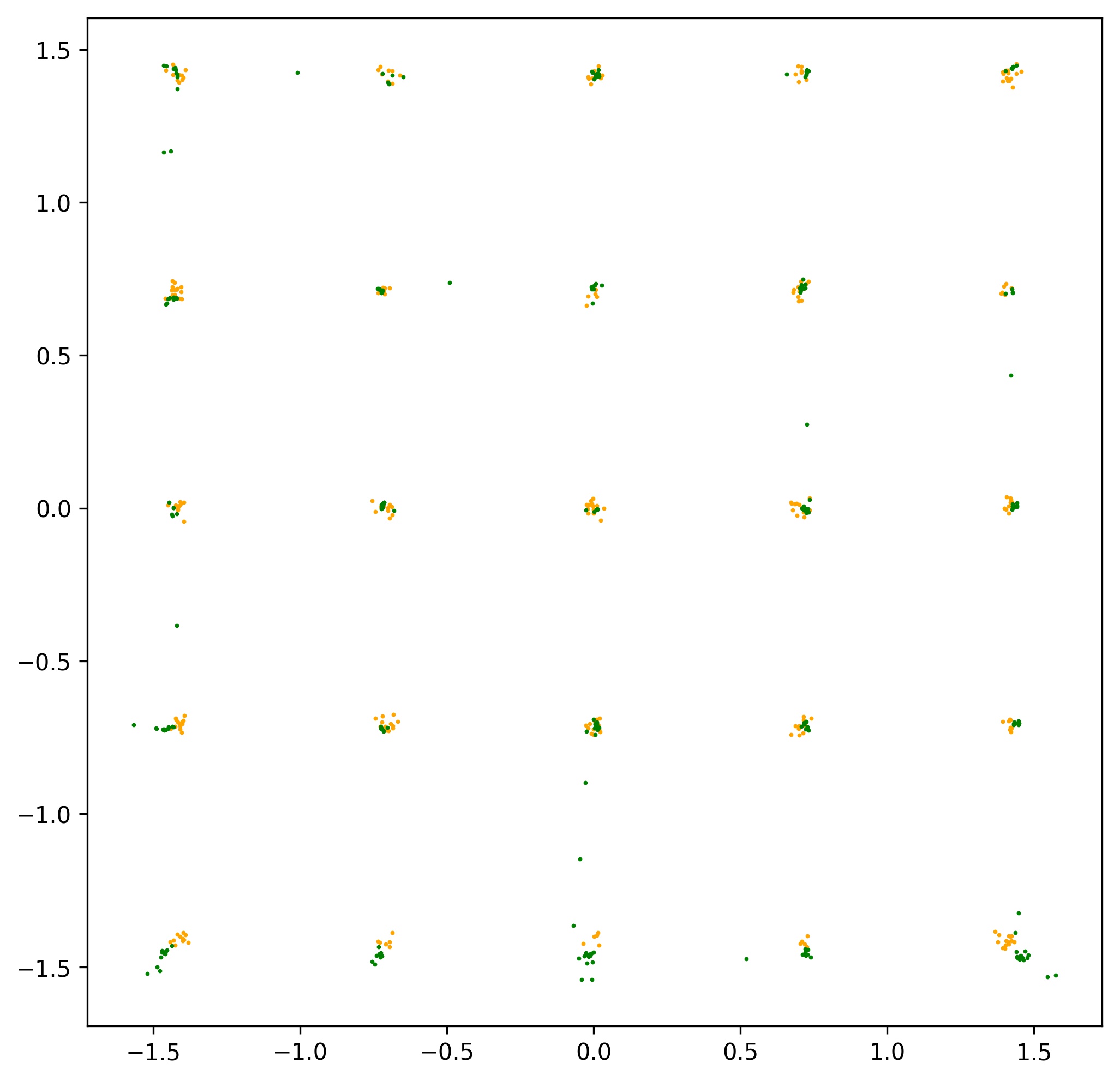}
		\end{minipage}
		&
		\hspace{-14pt}
		\begin{minipage}{0.2\linewidth}
			\includegraphics[width=\textwidth]{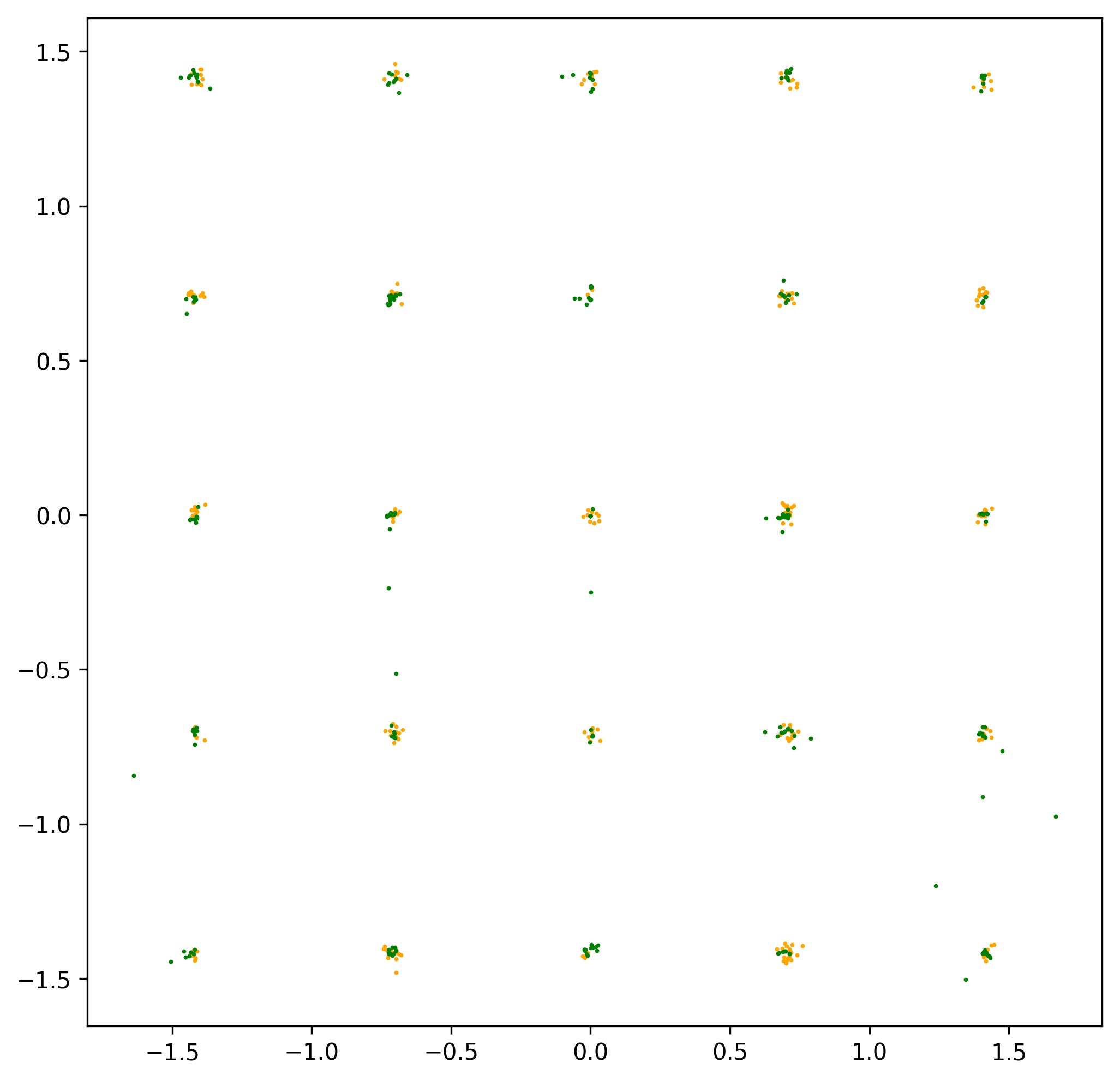}
		\end{minipage}
		&
		\hspace{-14pt}
		
		\begin{minipage}{0.2\linewidth}
			\includegraphics[width=\textwidth]{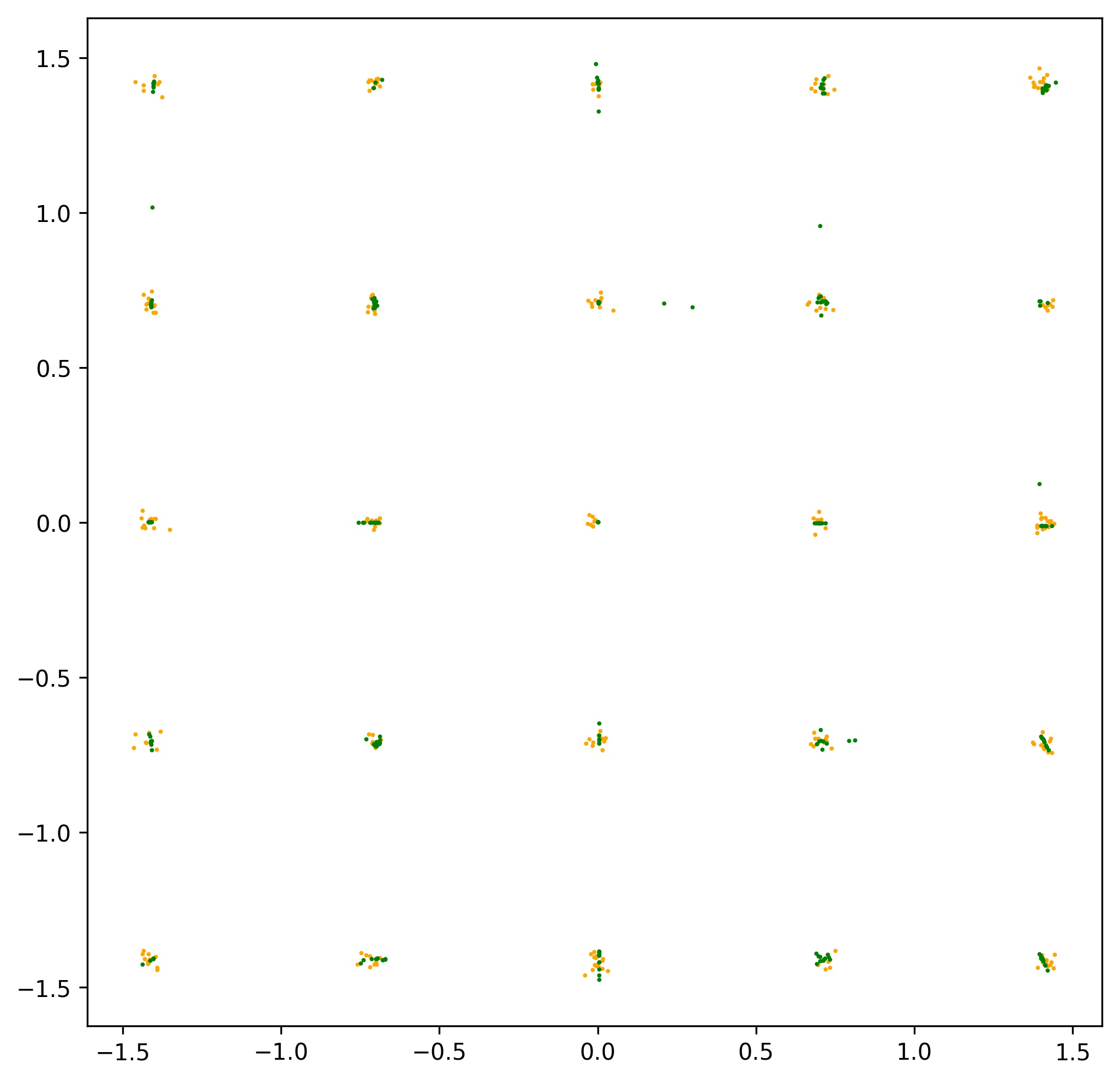}
		\end{minipage}
		\vspace{2pt}
		\\
		\textbf{Swiss Roll} &&&
		\vspace{1pt}
		\\
		\hspace{-8pt}
		\begin{minipage}{0.2\linewidth}
			\includegraphics[width=\textwidth]{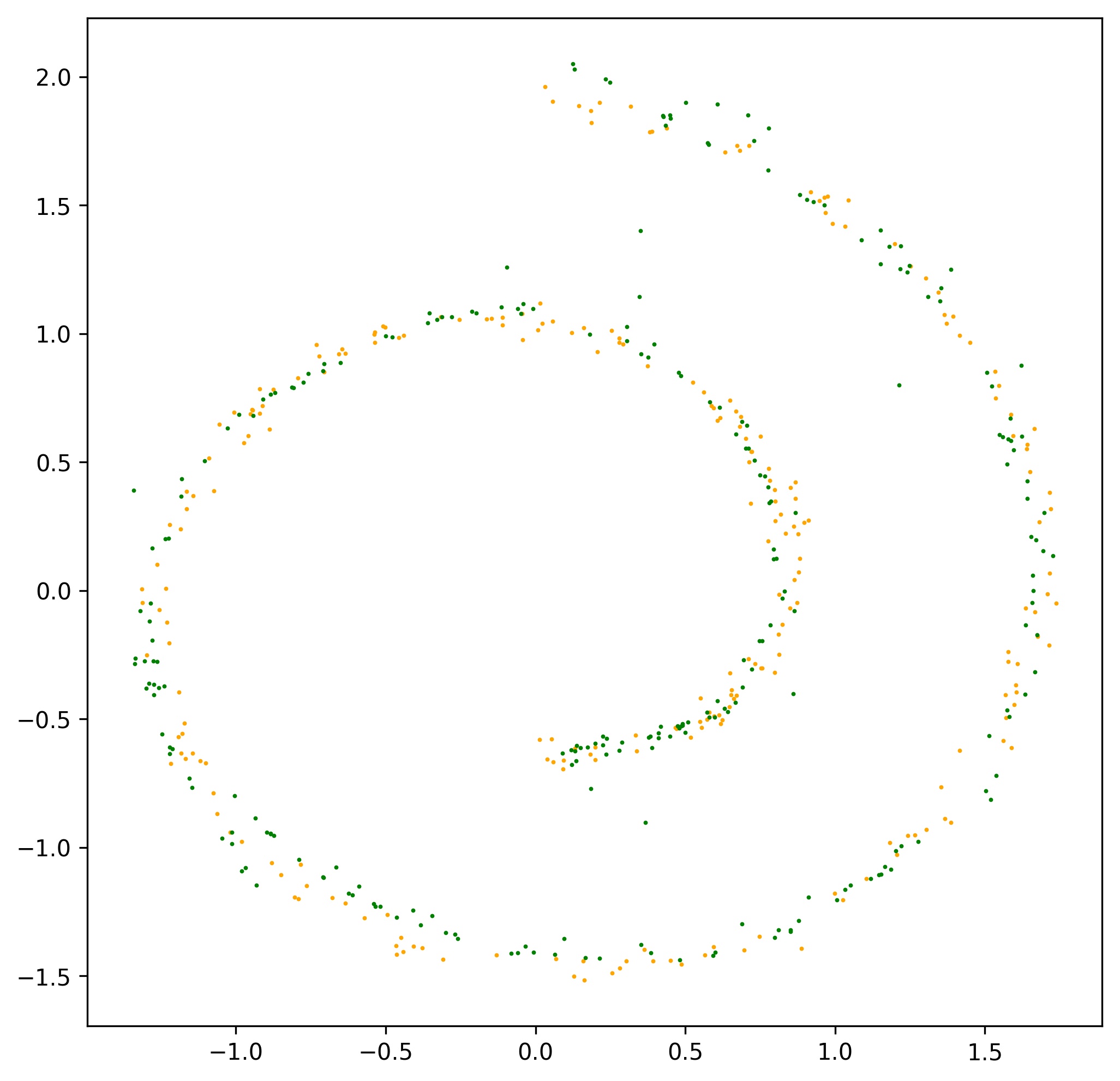}
		\end{minipage}
		&
		\hspace{-14pt}
		\begin{minipage}{0.2\linewidth}
			\includegraphics[width=\textwidth]{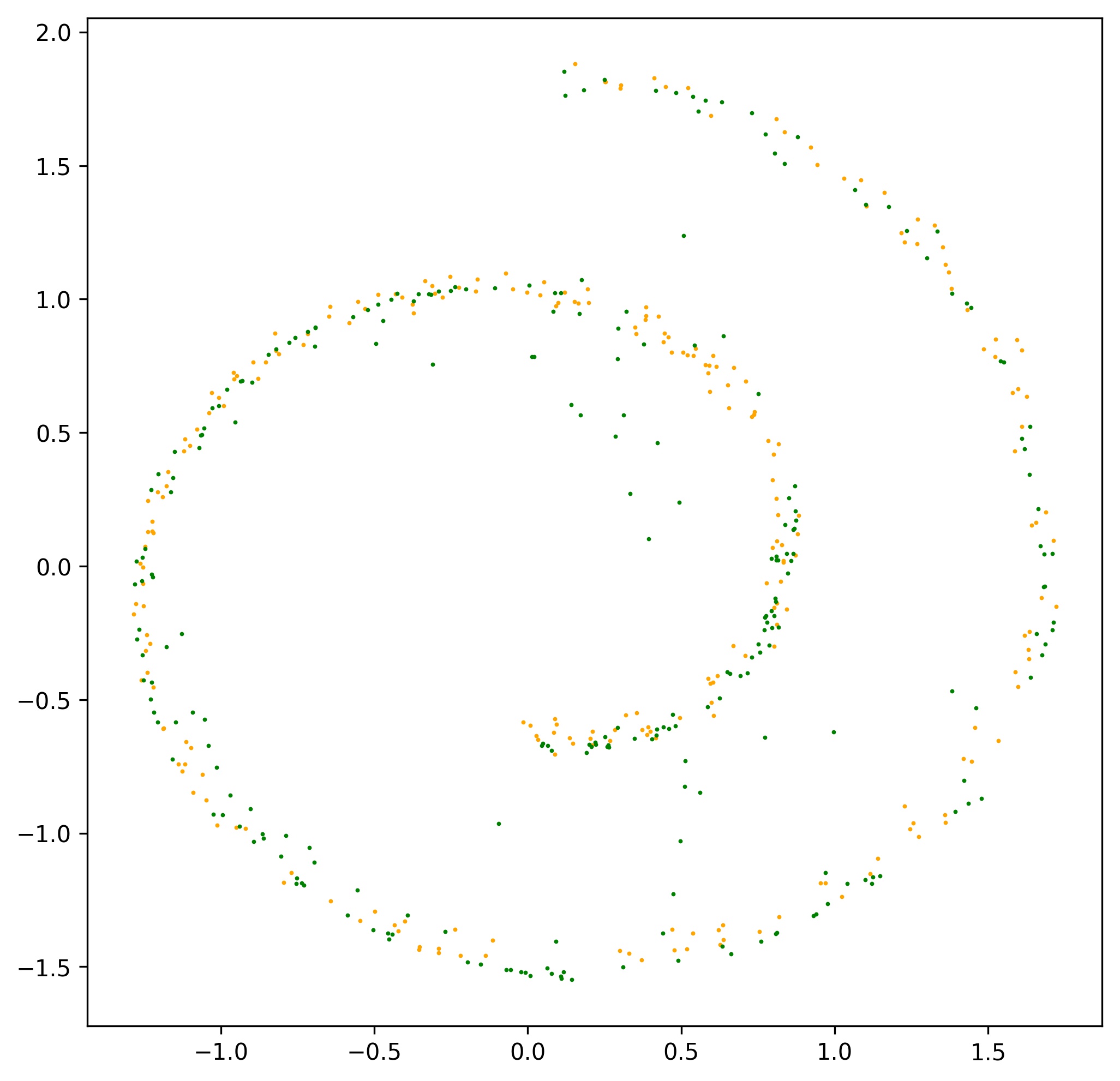}
		\end{minipage}
		&
		\hspace{-14pt}
		\begin{minipage}{0.2\linewidth}
			\includegraphics[width=\textwidth]{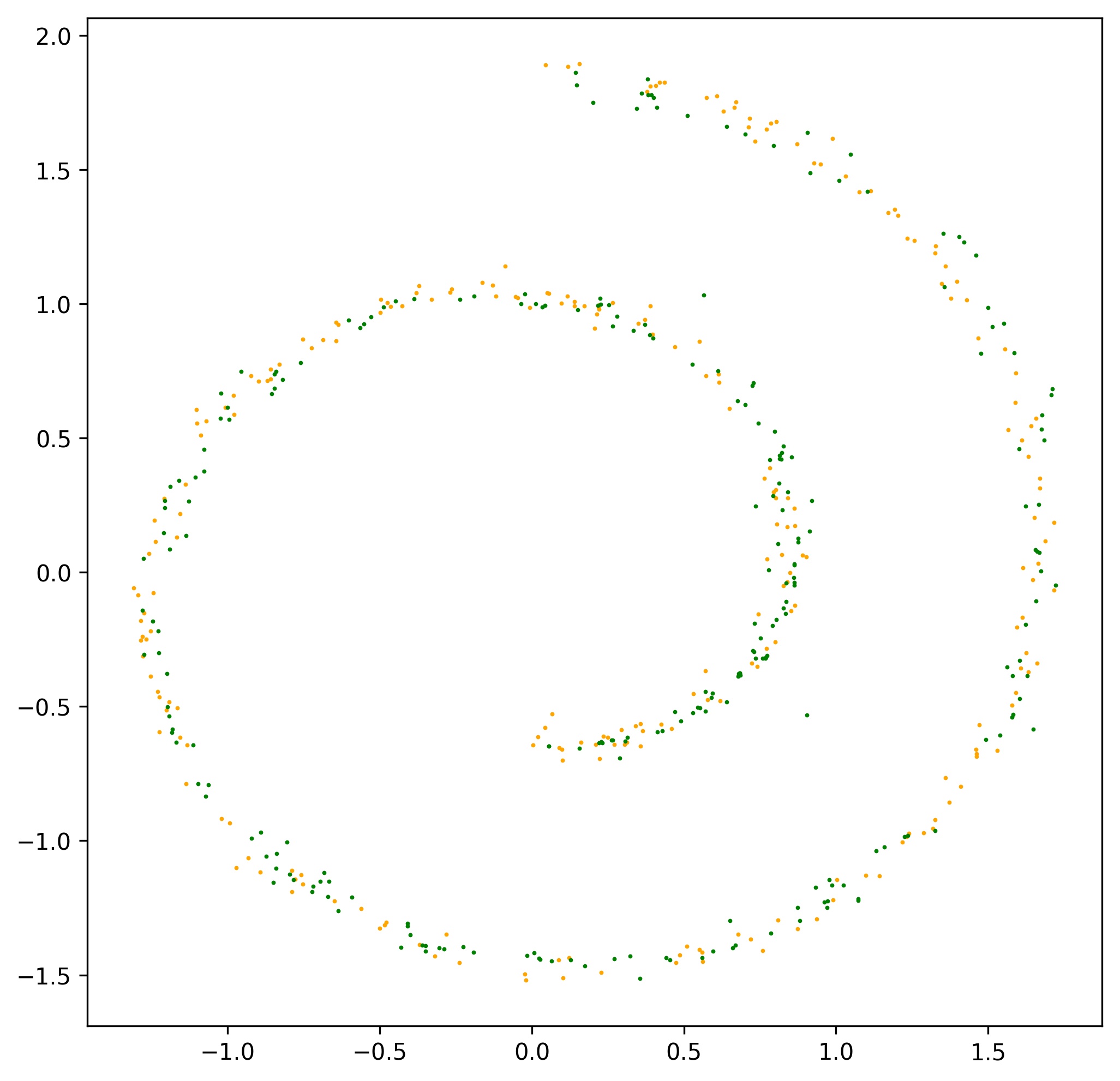}
		\end{minipage}
		&
		\hspace{-14pt}
		\begin{minipage}{0.2\linewidth}
			\includegraphics[width=\textwidth]{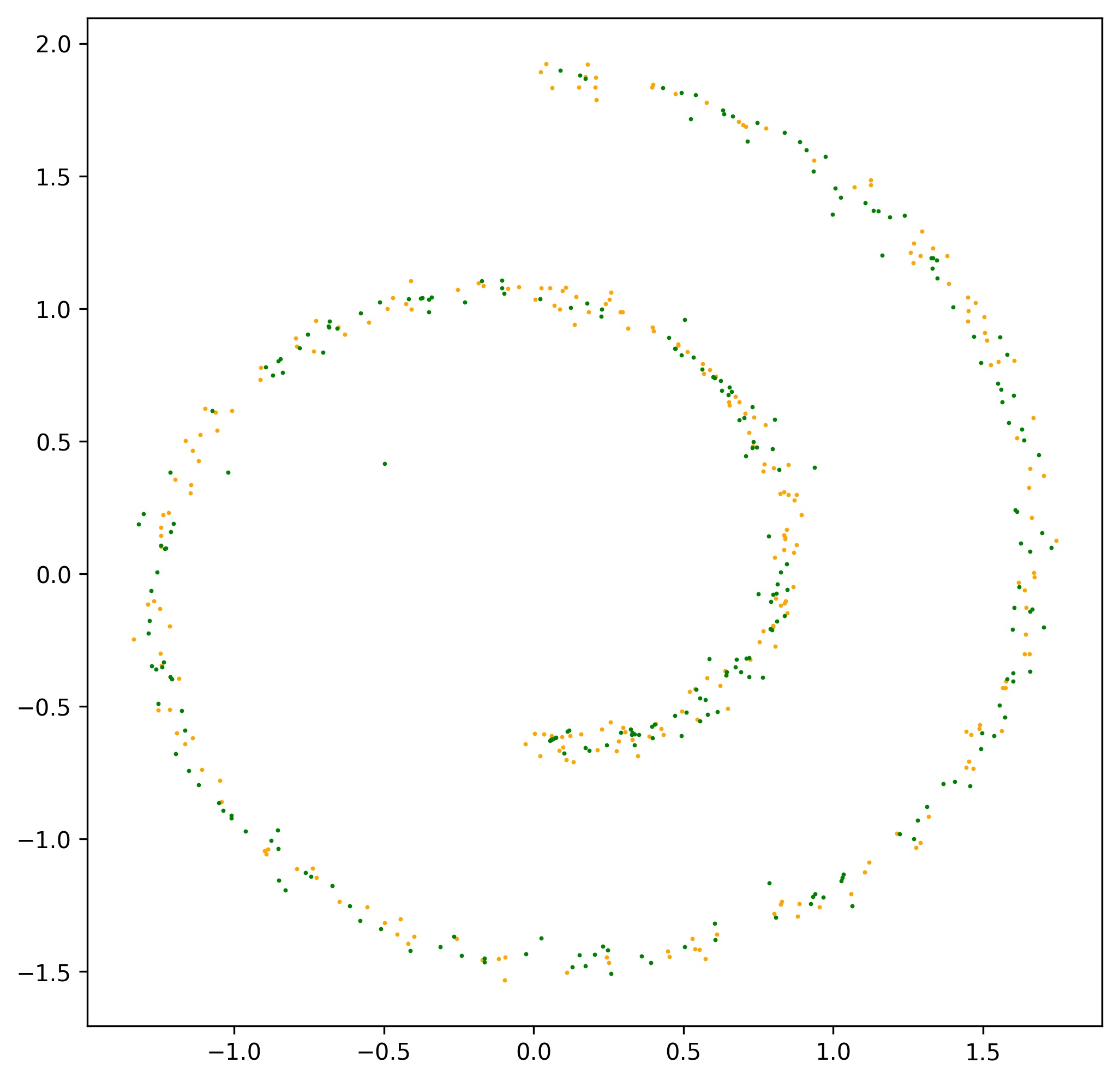}
		\end{minipage}
		\\
	\end{tabular}
	\caption{Comparing the performance of different models on toy datasets. The orange points show the samples from real data distribution $\mathbb{P}_r$, while the green points represent the samples from generated data distribution $\mathbb{P}_g$.}
	\label{toy}
\end{figure*}
\begin{equation}
\min_\theta \max_{w}  L_{\text{MMD}}\left(F_w(\mathbb{P}_r), F_w(\mathbb{P}_\theta)\right) + \lambda \text{Reg}
\end{equation}
where $\text{Reg}$ represents regularization terms (gradient penalty, $L1$ or $L2$ regularization). The training procedure for GAMN is described in Algorithm \ref{alg}. In the algorithm, $F_w(X):=\{F_w(x_i)\}_{i=1}^m, F_w(Y):=\{F_w(y_i)\}_{i=1}^m$, and see Equation (\ref{estimator}) for the definition of $\hat{L}_{\text{MMD}}(\cdot, \cdot)$ where a mixture of Gaussian kernels $k_{\bm{\sigma}}$ has been used. Note that $F_w$ is a CNN for image generation and MLP for toy data generation (e.g., a mixture of $8$ Gaussians, a mixture of $25$ Gaussians, and Swiss Roll) in this paper.

There is another thing to note. \citeauthor{ramdas2014high} show that the number of samples needed grows at least linearly with the number of dimensions to make Gaussian MMD test reliable. In GAMN, we use a mapper $F_w$ to map the origin high-dimensional image distribution to a low-dimensional feature representation distribution. Hence, we can significantly reduce the batch size.

\section{Experiments}

We carry out a series of experiments on low-dimensional and high-dimensional data to evaluate the competence of GAMN. The low-dimensional data consist of three toy datasets  \cite{gulrajani2017improved}, while the high-dimensional data include MNIST\cite{lecun1998gradient}, CIFAR-10 \cite{krizhevsky2009learning} and LSUN-Bedrooms dataset \cite{yu2015lsun}. We compare GAMN with notable MMD based models (GMMN and GMMN+AE) and state-of-the-art GAN based models (WGAN  and improved WGAN). To make the comparison fair, equivalent network architectures ($4$-layer $512$-dim ReLU MLP for low-dimensional data and DCGAN architecture \cite{radford2015unsupervised} for high-dimensional data) are used, except that GAMN has a $10$-dimensional output layer in the mapper. Considering that GMMN and GMMN+AE have different frameworks compared to GANs, we set them up based on the network architectures with default hyperparameters proposed in the original paper \cite{li2015generative} instead of the DCGAN architecture.

We use batch normalization \cite{ioffe2015batch} in the generator and layer normalization \cite{ba2016layer} in the mapper to stabilize GAMN's training procedure. For other models, we use the default setting in the original papers. We train GAMN with Adam optimizer, and other models are trained with the optimizers suggested in the original papers. Besides, batch size is set to $1000$ for GMMN and GMMN+AE according to the original paper \cite{li2015generative}. For WGAN, improved WGAN and GAMN, we follow the default settings of batch size in \cite{gulrajani2017improved}, i.e., $256$ on toy datasets, $50$ on MNIST, and $64$ on CIFAR-10 and LSUN-Bedrooms dataset. In this section, all the generated samples shown in the displayed figures are not cherry-picked.

\subsection{Performance on low-dimensional data}
To evaluate whether GAMN successfully learns the underlying data distribution $\mathbb{P}_r$, we first run experiments on toy datasets. Figure \ref{toy} shows samples drawn from different approaches on three toy datasets, whose ground truth distributions are a mixture of $8$ Gaussians, a mixture of $25$ Gaussians, and Swiss Roll, respectively. The figure illustrates that
WGAN, improved WGAN, GMMN and GAMN all achieve desirable performance on low-dimensional data. However, by careful examination of the generated samples, we observe that WGAN and improved WGAN do not work well on some datasets. For example, on 25 Gaussians data, WGAN fails to capture the underlying distributions accurately, and on Swiss Roll data, improved WGAN produces more outliers than GAMN and GMMN. Intriguingly, we also find that GMMN and GAMN almost learn the completely true distributions on all the datasets. In addition, we calculate MMD between $\mathbb{P}_r$ and $\mathbb{P}_g$ to show the degree of disagreement between the generated data distribution and real data distribution. The MMDs of different models are reported in Table \ref{MMDs}. Indeed, GAMN and GMMN significantly outperform WGAN and improved WGAN. For GAMN, $G$ minimizes $L_{\text{MMD}}\left(F(\mathbb{P}_r), F(\mathbb{P}_g)\right)$ but can achieve the lowest $L_{\text{MMD}}\left(\mathbb{P}_r, \mathbb{P}_g\right)$, which implies $F$ can assist $G$ to learn the underlying distribution. The superiority of GAMN and GMMN on toy datasets implies that MMD can serve as a reasonable and useful metric for distribution comparison, which provides a promising way for generative models to learn real data distribution. To further demonstrate this, we perform more experiments on high-dimensional data.

Note that we also train GMMN+AE on these datasets but it fails to generate reasonable samples. Thus we do not display the results here.

\begin{table}[htbp]
\centering
\begin{tabular}{llll}
\hline
 & 8G &25G & SR \\
\hline

WGAN & 0.300 & 0.312 & 0.305 \\
improved WGAN & 0.285 & 0.294 & 0.298\\
GMMN & 0.281 & 0.291 & 0.289\\
GAMN & \textbf{0.276} & \textbf{0.284} & \textbf{0.286}\\
\hline
\end{tabular}
\caption{Average MMDs over the last 1000 iterations. 8G, 25G and SR represent a mixture of 8 Gaussians, a mixture of 25 Gaussians and Swiss Roll dataset, respectively.}
\label{MMDs}
\end{table}

\subsection{Performance on high-dimensional data}
\begin{figure}[tbp]
	\centering
	\begin{minipage}{0.49\linewidth}
		\includegraphics[width=\textwidth]{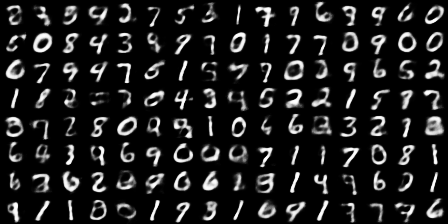}
	\end{minipage}
	\hspace{-4pt}
	\begin{minipage}{0.49\linewidth}
		\includegraphics[width=\textwidth]{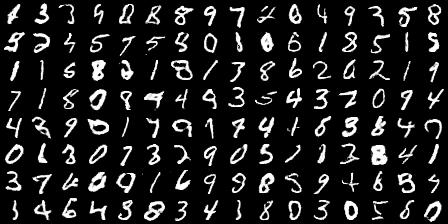}
	\end{minipage}
	\vfill
	\begin{minipage}{0.49\linewidth}
		\includegraphics[width=\textwidth]{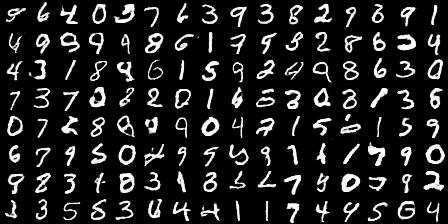}
	\end{minipage}
	\hspace{-4pt}
	\begin{minipage}{0.49\linewidth}
		\includegraphics[width=\textwidth]{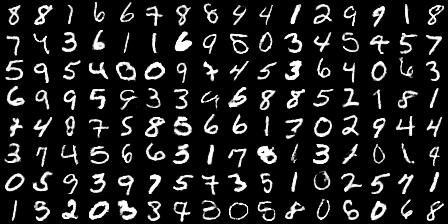}
	\end{minipage}
	\caption{ Visualization of the generated samples on MNIST using GMMN+AE (top-left), GAMN (top-right), WGAN (bottom-left) and improved WGAN (bottom-right). The samples are 28x28 images with 1 color channel.}
	\label{mnist}
\end{figure}

\begin{figure}[!h]
	\centering
	
	\begin{minipage}{0.95\linewidth}
		\includegraphics[width=\textwidth]{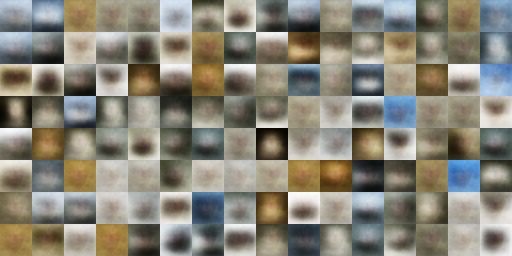}
	\end{minipage}
	
	\vspace{1pt}
	\begin{minipage}{0.95\linewidth}
		\includegraphics[width=\textwidth]{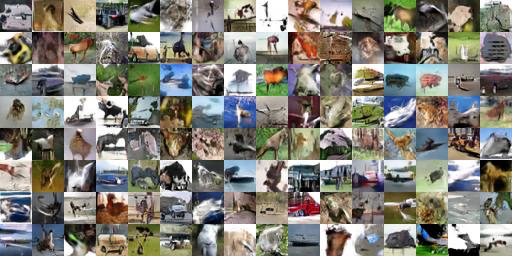}
	\end{minipage}
	
	\vspace{1pt}
	\begin{minipage}{0.95\linewidth}
		\includegraphics[width=\textwidth]{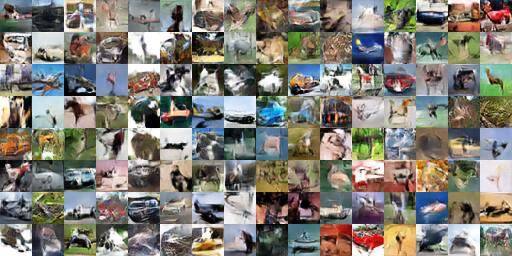}
	\end{minipage}
	
	\vspace{1pt}
	\begin{minipage}{0.95\linewidth}
		\includegraphics[width=\textwidth]{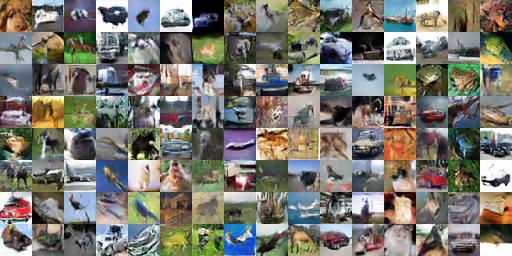}
	\end{minipage}
	\caption{Visualization of the generated samples on CIFAR-10 using GMMN+AE, GAMN, WGAN and improved WGAN (from top to bottom), repectively. The samples are 32x32 images with 3 color channels.}
	\label{cifar}
\end{figure}

Digital images can be regarded as points in a high-dimensional space, in which each dimension corresponds to the chroma of every pixel. In the high-dimensional space, images lie on the complicated geometry of manifolds, which makes it a challenge for generative models. To validate our performance on high-dimensional data, we train GAMN on MNIST, CIFAR-10 and LSUN-Bedrooms dataset and compare the generated results with GMMN, GMMN+AE, WGAN and improved WGAN. Figure \ref{mnist} and \ref{cifar} show the generated samples on MNIST and CIFAR-10.

\begin{table}[tbp]
	\begin{minipage}{0.48\linewidth}
		\begin{tabular}{ll}
			\hline
			Method & Score \\
			\hline
			GMMN & 2.45\\
			GMMN+AE & 2.78\\
			WGAN & 5.61 \\
			imp WGAN & 6.16 \\
			GAMN & \textbf{6.44}\\
			\hline
		\end{tabular}
		\caption{Inception scores. (imp WGAN refers to improved WGAN)}
		\label{incs}
	\end{minipage}
	\hfill
	\begin{minipage}{0.48\linewidth}
		\begin{tabular}{lll}
			\hline
			$d$ & Reg  & Score \\
			\hline
			10 & GP & \textbf{6.44} \\
			2 & GP & 6.43\\
			3 & GP  & 6.43\\
			10 & L1  & 6.35\\
			10 & L2  & 6.25\\
			10 & classical L2  &6.01\\
			\hline
			
		\end{tabular}
		\caption{Inception scores of GAMN under different conditions.}
		\label{rob}
		
	\end{minipage}
\end{table}
On the one hand, we compare GAMN with the different MMD based models, i.e., GMMN, GMMN+AE. Considering that GMMN+AE is superior to GMMN on complex high-dimensional data \cite{li2015generative}, we only show GMMN+AE here for comparison. In fact, we also run GMMN and get results slightly worse than GMMN+AE (see Supplementary Figure 1), which indeed indicates that direct optimization of MMD does not work well in complicated high-dimensional space. We carefully compare the generated samples from GMMN+AE and GAMN. It is not hard to see that GAMN significantly outperforms GMMN+AE. In Figure \ref{mnist}, GMMN+AE generates a batch of fuzzy digits whose sharpness is much lower than GAMN does. More obviously in Figure \ref{cifar}, GMMN+AE fails to generator meaningful samples while GAMN can successfully produce vivid images. Two reasons are speculated to account for the dramatical superiority of GAMN.  First, the mapper in GAMN, which is dynamically fine-tuned during an adversarial training process, greatly outperforms the auto-encoder in GMMN+AE in assisting the generator to optimize MMD in high-dimensional space. Second, the convolutional layers in the mapper can learn hierarchical feature representations for images.

\begin{figure*}[!h]
	\centering
	
	\begin{minipage}{0.33\linewidth}
		\includegraphics[width=\textwidth]{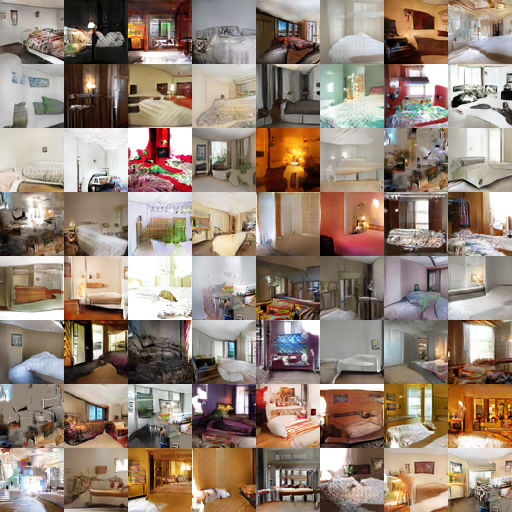}
	\end{minipage}
	\begin{minipage}{0.33\linewidth}
		\includegraphics[width=\textwidth]{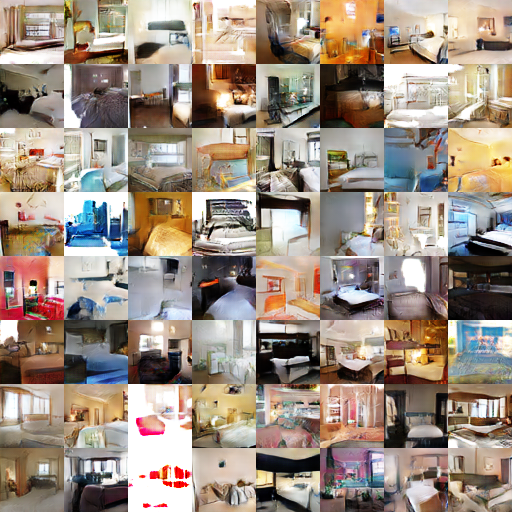}
	\end{minipage}
	\begin{minipage}{0.33\linewidth}
		\includegraphics[width=\textwidth]{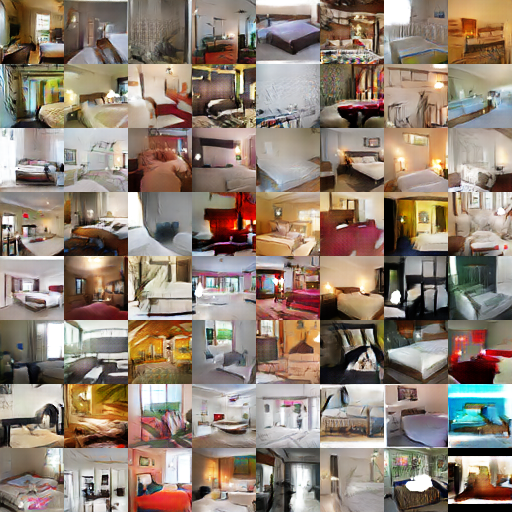}
	\end{minipage}
	\caption{Visualization of the generated samples on LSUN-Bedrooms dataset using GAMN (left), WGAN (middle) and improved WGAN (right). The samples are 64x64 images with 3 color channels.}
	\label{lsun}
\end{figure*}

On the other hand, GAMN achieves competitive performance with WGAN and improved WGAN in terms of image quality and diversity on both MNIST and CIFAR-10 datasets. To further compare the performance on larger images,
we train GAMN, WGAN, and improved WGAN on LSUN-Bedrooms dataset. As shown in Figure \ref{lsun}, GAMN produces lifelike bedroom images with smooth brush strokes, meticulous texture, and soft colors. Visually, the sample quality of GAMN is better than WGAN and comparable with improved WGAN.

\begin{figure}[htbp]
	\begin{minipage}{0.49\linewidth}
		\includegraphics[width=\textwidth]{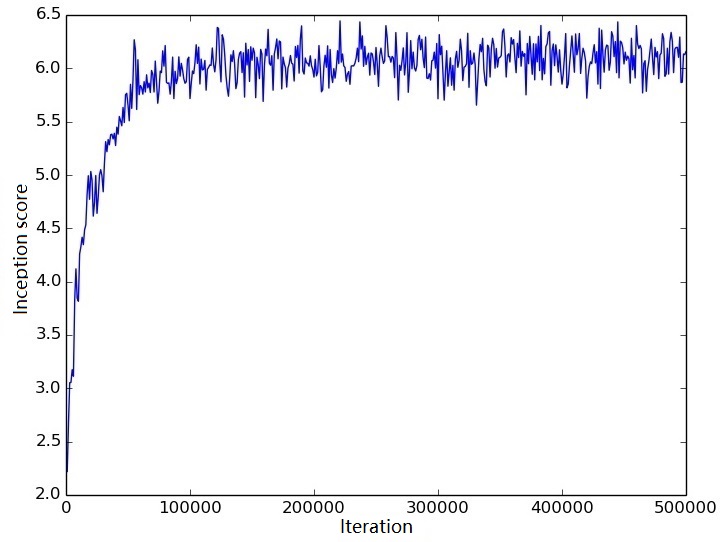}
	\end{minipage}
	\begin{minipage}{0.49\linewidth}
		\includegraphics[width=\textwidth]{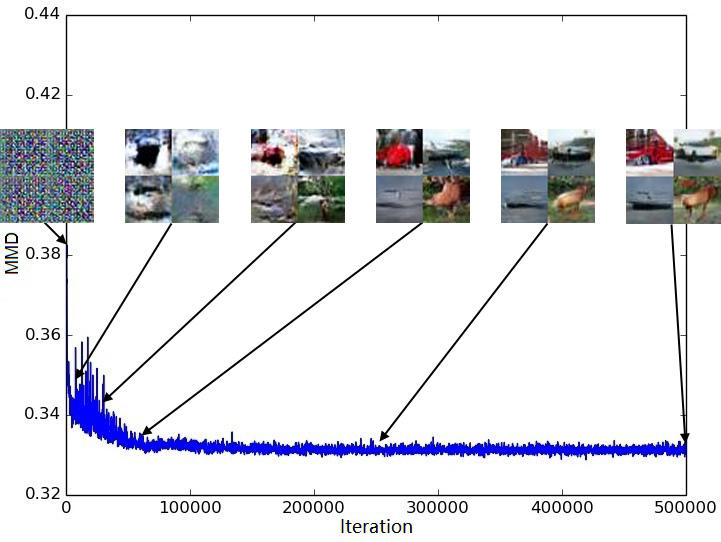}
	\end{minipage}
	\caption{Inception score and MMD over iterations during GAMN's training on CIFAR-10. MMD decreases consistently as training progress and in the meantime inception score and sample quality increase.}
	\label{plot}
\end{figure}

In addition to the visual comparison, we also perform a quantitative assessment. Inception score \cite{salimans2016improved} has been widely used to measure the image quality of generated samples quantitatively. Thus, we compare the best inception scores of different models on CIFAR-10 dataset. As shown in Table \ref{incs}, GAMN achieves significantly superior to GMMN and GMMN+AE, and slightly better performance than WGAN and improved WGAN. Figure \ref{plot} plots the inception scores and MMD over iterations during GAMN's training on CIFAR-10. It demonstrates that inception score, image quality, and MMD agree well with each other. This property implies that MMD can also serve as a reasonable metric for sample quality evaluation, which is useful for monitoring of training process and model comparison. What is different from Wasserstein distance in WGAN and improved WGAN is that MMD is independent to the critics (i.e., the discriminators in GAN based models and the mapper in GAMN), which provides a convenient way to compare models with different critics.

\subsection{Robustness}\label{method}
To demonstrate the robustness of our model, we train GAMN on CIFAR-10 with various mapping dimensions $d$ (i.e., the dimension of feature representation space $\mathcal{R}$) and regularization terms (i.e., gradient penalty, L1 regularization, L2 regularization and classical L2 regularization\footnote{Classical L2 regularization here refers to L2 regularization on all the parameters of the mapper $F$.}). We find that GAMN with different settings can all achieve reasonable inception scores (Table \ref{rob}) and generate high-quality images (see Supplementary Figure 2). It illustrates that GAMN still works well even with low mapping dimension or naive regularizations. It should be noted that the using L2 regularizatin on the normalization layer of the mapper alone is slightly better than the classical L2 regularization. In addition, we empirically recommend adding an MMD term $L_{\text{MMD}}(\mathbb{P}_r, \mathbb{P}_g)$ in the loss function of the generator when using gradient penalty to acquire a slight improvement of performance.

\section{Related Work}
This paper proposes a novel generative model, called GAMN, which integrates GAN framework with GMMN. Here we discuss how it relates to existing works.

\citeauthor{goodfellow2014generative}  firstly propose an adversarial game  framework to train generative models which they name generative adversarial networks (GANs). GANs contain a generator and a discriminator. The generator is fed with a random vector sampled from a prior to produce fake samples which look similar to the samples in a given dataset. The discriminator is trained to distinguish real data samples from generated fake samples while the generator is trained to confuse the discriminator. The competition between the generator and discriminator helps the generator to model the underlying distribution of the dataset better. After that, \citeauthor{radford2015unsupervised} introduce CNN to the GAN framework and explore a family of architectures called DCGANs which make training higher resolution and deeper generative models possible.

$f-$GAN is proposed to minimize the variational lower bound on $f-$divengence between two distributions \cite{nowozin2016f}. EBGAN does total variation distance minimization to learn the underlying distribution \cite{zhao2016energy,arjovsky2017wasserstein}. However, these GAN based models are still hard to train, which need a careful balance during the adversarial optimization. WGAN and improved WGAN are proposed to address  this issue, which are trained to minimize Wasserstein divergence \cite{arjovsky2017wasserstein,gulrajani2017improved}.

More related to our method, generative moment matching networks GMMNs \cite{li2015generative,dziugaite2015training} optimize MMD between distributions to learn generative models from data. The original GMMN only includes a single generator. To boost the performance of GMMN, \citeauthor{li2015generative} introduce an auto-encoder to a GMMN and name it GMMN+AE. They first use an auto-encoder to produce the code representations of data and then minimize MMD between data code distribution and generated code distribution. The functionality of the auto-encoder here is similar to that of the mapper in GAMN to some extent: both of them map the data distribution from the real data space to another code (representation) space and can reduce the dimension of high-dimensional images. However, the mapper in GAMN is entirely different from the auto-encoder in GMMN+AE. The most different part is that the mapper is a dynamic network which always changes during training process to be a good adversary for the generator, while the auto-encoder is a static network which is trained in the beginning and then keeps fixed all the time. We believe this difference makes the mapper much powerful than the auto-encoder, thus enhancing the generator in GAMN considerably. Our experiments also demonstrate this.

In a very recent independent work (uploaded on arXiv on 24 May 2017), \citeauthor{li2017mmd} propose a new method, called adversarial kernel learning, which is very similar to ours. They require the kernel to satisfy some additional theoretical properties (such as being characteristic). Hence they need to train an extra auto-encoder to obtain certain nontrivial regularization term in the loss function.
Our model GAMN is much simpler and does not need such regularization terms,
yet can still generate very reasonable results.
In the experiments on CIFAR-10, the inception score of GAMN ($6.44$) is slightly higher than that of their model ($6.24$, as reported in their paper).
As we only discovered this work very recently,
we have to leave a more detailed comparison to the future work.

\section{Conclusions}
In this paper, we integrate GAN framework with MMD and propose a novel generative model called GAMN. We show that the adversarial mapper $F$ in GAMN can help the generator capture the underlying distribution of real data better and reduce the batch size needed for training considerably. We also demonstrate that GAMN performs significantly better than GMMN and GMMN+AE (two existing generative models that also use MMD), and slightly better or comparable with state-of-the-art GAN based methods on benchmark datasets with the same architecture.

There are many interesting directions for future research.  Firstly, the robustness of GAMN for training can allow us to explore a wider range of architectures like ResNet \cite{he2016deep} and DenseNet \cite{huang2016densely} as well as more complex kernels instead of Gaussian kernels to improve sample quality. Secondly, by adding AC-GAN conditioning \cite{odena2016conditional}, we can extend GAMN to a conditional generative model.

\bibliographystyle{aaai}
\bibliography{references}

\clearpage
\setcounter{figure}{0}
\renewcommand{\figurename}{Supplementary Figure}

\begin{figure*}[h!]
\centering

\begin{minipage}{0.47\linewidth}
  \includegraphics[width=\textwidth]{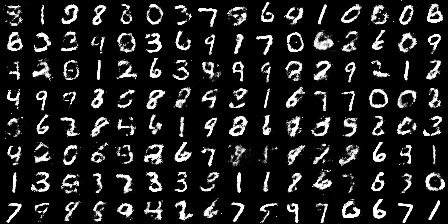}
\end{minipage}
\hspace{0pt}
\begin{minipage}{0.47\linewidth}
  \includegraphics[width=\textwidth]{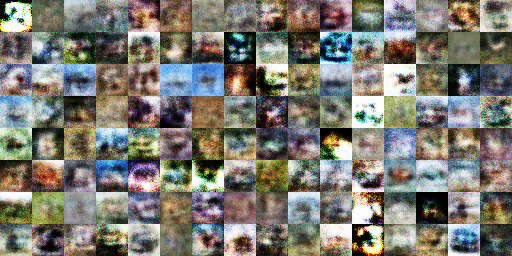}
\end{minipage}

\caption{Visualization of the generated samples on MNIST (left) and CIFAR-10 (right) using GMMN. }
\label{suppgmmn}
\end{figure*}

\begin{figure*}[h!]
\centering
\begin{minipage}{0.47\linewidth}
  \includegraphics[width=\textwidth]{gamn_cifar.jpg}
\end{minipage}
\hspace{0pt}
\begin{minipage}{0.47\linewidth}
  \includegraphics[width=\textwidth]{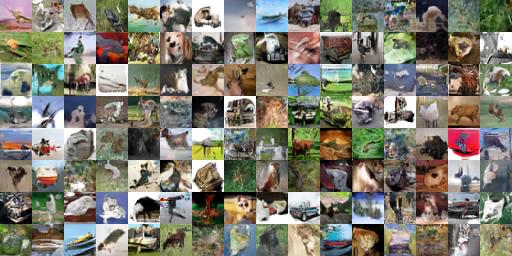}
\end{minipage}

\begin{minipage}{0.47\linewidth}
\vspace{4pt}
  \includegraphics[width=\textwidth]{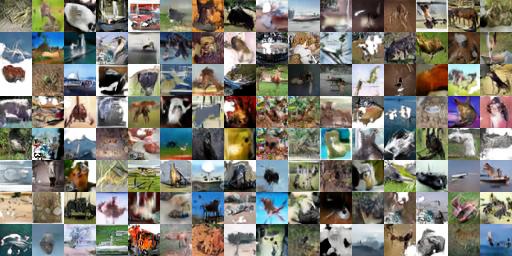}
\end{minipage}
\hspace{0pt}
\begin{minipage}{0.47\linewidth}
\vspace{4pt}
  \includegraphics[width=\textwidth]{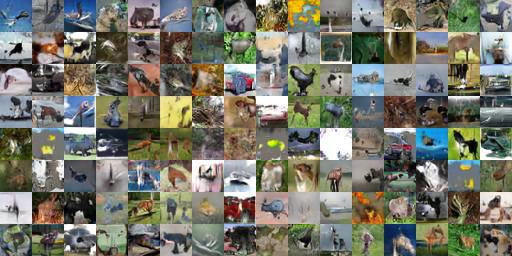}
\end{minipage}

\begin{minipage}{0.47\linewidth}
\vspace{4pt}
  \includegraphics[width=\textwidth]{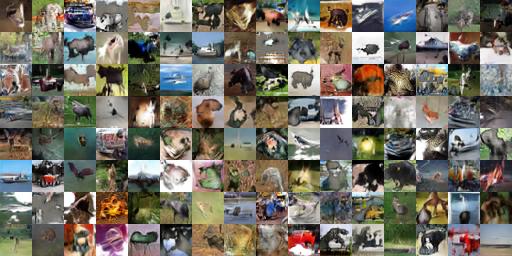}
\end{minipage}
\hspace{0pt}
\begin{minipage}{0.47\linewidth}
\vspace{4pt}
  \includegraphics[width=\textwidth]{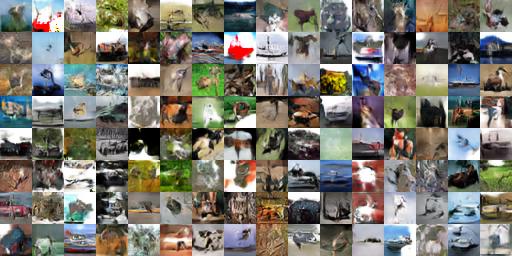}
\end{minipage}
\caption{Visualization of the generated samples on CIFAR-10 using GAMN under different conditions. (top-left) 10 mapping dimension, gradient penalty. (top-right) 2 mapping dimension, gradient penalty. (middle-left) 3 mapping dimension, gradient penalty. (middle-right) 10 mapping dimension, L1 norm. (bottom-left) 10 mapping dimension, L2 norm. (bottom-right) 10 mapping dimension, classical L2 norm.}
\label{suppparas}
\end{figure*}

\end{document}